\pdfoutput=1

\documentclass[11pt]{article}

\usepackage[final]{acl}

\usepackage{times}
\usepackage{latexsym}
\usepackage{fontawesome}
\usepackage[T1]{fontenc}
\usepackage{listings}

\usepackage[utf8]{inputenc}
\usepackage[inline]{enumitem}
\usepackage{tcolorbox} 
\usepackage[normalem]{ulem}

\usepackage{microtype}

\usepackage{inconsolata}
\usepackage{amsmath}
\usepackage{booktabs}
\usepackage{multirow}
\usepackage{multicol}
\usepackage{graphicx}

\usepackage[dvipsnames]{xcolor} 
\usepackage{siunitx} 

\newcommand{\ann}[2]{\textcolor{#1}{\textbf{#2}}}
\title{%
   Auto-AdvER: Disentangling Condition, History, and Commercial Offers in Automotive Advertisements   \; \includegraphics[scale=0.08]{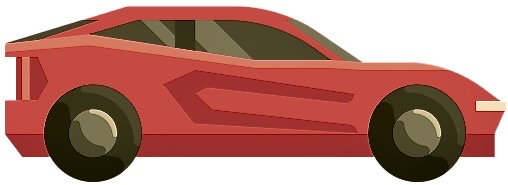}
}

\author{
Filippos Ventirozos\textsuperscript{1,2},
Ioanna Nteka\textsuperscript{2},
\textbf{Tania Nandy\textsuperscript{2}},
\textbf{Jozef Ba\v{c}a\textsuperscript{2}},\\
\textbf{Charlie Roadhouse\textsuperscript{1}},
\textbf{Peter A. Appleby\textsuperscript{2}} \and
\textbf{Matthew Shardlow\textsuperscript{1}} \\
\textsuperscript{1}\,Manchester Metropolitan University, UK \\
\textsuperscript{2}\,Autotrader Research Group, Autotrader UK \\
 \texttt{\{\href{mailto:f.ventirozos@mmu.ac.uk}{f.ventirozos}, \href{mailto:m.shardlow@mmu.ac.uk}{m.shardlow}\}@mmu.ac.uk}
}

\begin{document}
\maketitle
\begin{abstract}

This paper presents a case study on the development of \textbf{Auto-AdvER}, a specialised span-labelling schema, along with its corresponding dataset for the textual car advertisement domain. Developed with industry needs in mind, \textbf{Auto-AdvER} is designed to enhance text mining analytics in this domain and contributes a linguistically unique sequence-labelling dataset\footnote{The annotation schema and dataset can be found at \url{https://github.com/FilipposVentirozos/Auto-AdvER}}. We introduce a schema consisting of three labels: \textit{`Condition'}, \textit{`Historic'}, and \textit{`Sales Options'}. We outline the guiding principles for annotation, describe the methodology for schema development, and present the results of an annotation study demonstrating an inter-annotator agreement of 78\%. Furthermore, we compare the performance of various models, including transformer encoder models (BERT, DeBERTaV3) and transformer decoder-based open- and closed-source large language models: Llama, Qwen, GPT, and Gemini, as well as conditional random fields. Our results show that a fine-tuned DeBERTaV3 encoder achieves the highest $F_1$-score, followed closely by conditional random fields, while the large language models reach the highest precision but lower recall. We present this work as a stepping stone towards more fine-grained analysis and discuss \textbf{Auto-AdvER}'s potential impact on advertisement analytics, pricing, and customer insights. Potential applications include the analysis of market dynamics and data-driven predictive maintenance.


\end{abstract}



\section{Introduction}

The automotive industry relies heavily on accurate data about vehicles that are manufactured, sold, and resold. For digital businesses, much of this information is contained in free-text online car advertisements, where it is used for applications such as market analysis, competitive intelligence, and customer insights. Converting these descriptions into structured representations is therefore an important domain-specific information extraction problem. In this domain, the relevant text spans extend beyond classical named entities to include attribute-value and condition expressions. We therefore frame the task as span-based extraction from automobile advertisements, operationalised using a named entity recognition (NER)-style sequence-labelling approach \citep{li2020survey}. Many of the spans we target (e.g., ``smoke free'', ``tyre tread is low'') are descriptive attributes or keyphrase-like mentions rather than names; we therefore refer to the annotated units as \emph{spans} and to the three categories as \emph{span labels}, and use NER terminology only for the modelling framework and for prior work. To support this task, we introduce Auto-AdvER (\textbf{Auto}mobile \textbf{Adv}ertisement \textbf{E}ntity \textbf{R}ecognition), a novel industry-developed annotation schema for vehicle-related spans in advertisement text.


Alongside the schema, we set out the principles we followed in forming its three labels (\textit{Condition}, \textit{Historic}, and \textit{Sales Options}) and in deciding which words each label covers. We hope \textbf{Auto-AdvER} sets a standard for text annotation in the automotive industry and elsewhere. We make three contributions:

\begin{enumerate}[leftmargin=*,itemsep=0pt,topsep=2pt]
    \item A novel domain span labelling dataset
    \item The comparison of several applied span labelling models
    \item Critical discussion of these results and applicability to the automotive industry and beyond.
\end{enumerate}

The paper is structured as follows: Section~\ref{sec:rel-work} introduces related information extraction studies in the automotive domain. In Section~\ref{sec:ner-schema}, we describe our annotation guidelines and methodology, and in Section~\ref{sec:data_attrs} we present an overview of the dataset. Section~\ref{sec:exp} presents our experiments and inter-annotator agreement (IAA) results. We discuss wider applicability and future work in Section~\ref{sec:discussion}, and conclude in Section~\ref{sec:concl}.

\section{Related Work}
\label{sec:rel-work}

Sequence labelling is a foundational modelling framework for information extraction in NLP: every token receives a label, and runs of labelled tokens form the extracted units. NER is its most prominent instance, with early datasets developed for the CoNLL evaluation campaigns \cite{tjong-kim-sang-de-meulder-2003-introduction} focusing on general news text. NER involves identifying spans in text applicable for labelling and classifying them into predefined categories --- a process often performed end-to-end \citep{JEHANGIR2023100017}. Typical approaches have employed sequence labelling models such as conditional random fields \citep{keraghel2024recentadvancesnamedentity}, BiLSTMs \citep{chiu-nichols-2016-named} or transformers \cite{hanh-2021-named}.

NER, in turn, has been extensively explored across various domains, often serving as the initial step in information extraction projects. Domain-specific datasets and models have been developed for biomedicine \citep{fu-etal-2023-biomedical}, agriculture \citep{G2023120440}, and music \citep{epure-hennequin-2023-human}. In each case, a bespoke annotation schema had to be developed to represent the units of interest in that domain. Our work follows the same path, but the units we annotate are descriptive spans rather than names, as discussed below. 

There has been a growing body of literature in information extraction for the automotive domain. To the best of our knowledge, the earliest relevant work is by \citet{lothritz-etal-2020-evaluating}, who evaluate NER models on the English/Turkish Wikipedia dataset of \citet{sahin2017wikipedia}, whose articles span multiple domains, including the automotive domain. The annotation tags included automotive-specific entities such as fuel, engine, dates, manufacturers, models, generation and vehicle types. \citet{park2023admit} introduced ADMit, which combines adversarial training and multi-task learning for automotive NER in Korean and English. Their work addresses domain adaptation between general-domain and automotive-specific terminology, but focuses on FAQ systems for automotive conversational AI assistants. \citet{10499162} developed methods for detecting automotive parts/accessories in Chinese. FindVehicle \citep{guan2024findvehicle} targets vehicle retrieval with entity types such as vehicle colour, brand, model, and location. 

In contrast, our work focuses on extracting descriptive spans concerning a car’s condition—e.g., scratches, malfunctioning components, tyre condition, and current paperwork status—as well as historical mentions and past events (e.g., accidents, component changes), and finally, commercial offers (e.g., warranties offered, part-exchange offers, delivery options), as specified in our annotation guidelines. Beyond advertisements, the same schema could be used for dealer listings, classifieds, and accident or insurance reports, enabling cross-regional analyses of commercial offerings and condition mentions for supporting safety and reliability monitoring.

\section{Dataset}
\label{sec:ner-schema}
\subsection{Annotation Methodology}
To our knowledge, this is the first schema for labelling car text advertisements. Consequently, we developed our labels by working closely with our industry partner, Autotrader UK, and observing processes similar to the action research framework \citep{Molineux2018UsingAR}, bridging industry and academia. Through our collaboration, we engaged extensively with industry data and resources, consulting with infrastructure engineers to gain an in-depth understanding of the automotive domain. This in-house expertise was pivotal in crafting our project's initial set of labels and guidelines. Below, we outline the core principles for developing our guidelines: \begin{enumerate}[leftmargin=*,itemsep=0pt,topsep=2pt]
\itemsep0em
 \item To construct the schema, we adopt a broad-category-first design: we delineate a small number of broad, semantically distinct categories and leave their refinement into finer-grained sub-types, and the linking of individual mentions to specific entities, to future work (Appendix~\ref{sec:appendix:labels} lists the sub-types we observed).


 \item Each label was designed to be self-descriptive and unambiguous, such that a human annotator could reliably extract the correct spans by asking a single, distinct question per label (e.g., ``What are the Condition mentions?'') in order to minimise ambiguous overlapping labels.

 \item  The aim was to annotate only factual statements, not opinions or personal sentiments. 
 

   
\end{enumerate}



\subsubsection{Iterations}
Our methodology for developing the guidelines involved two separate teams working in iterations: a university-based team (two members) and an industry-based team (three members). The university team would create or edit the classes and annotation guidelines. Then, these would subsequently be discussed with the industry-based team, which comprised domain-expert data scientists. We undertook ten iterations of annotation, with feedback as detailed in the following two sections.

\paragraph{Initial Iterations}

During the first set of iterations, the university team proposed a draft of the guidelines. These proposals were then refined by the industry team, focusing on two main aspects: \begin{enumerate}[leftmargin=*,itemsep=0pt,topsep=2pt]
\itemsep0em
    \item The inclusivity of each class and the extent to which they covered various aspects of the advertisements.
    \item The identification of any omitted classes, which could either be sourced from other data or deemed non-essential.
\end{enumerate}

\noindent

Following these discussions, the university team further refined the classes. When a class lacked clarity regarding its encompassed aspects, the team evaluated the semantic coherence of each aspect within the proposed definitions. Rapid prototyping enabled them to explore alternative classes and assess their detectability.

\paragraph{Secondary Iterations}

As the labels began to take shape, we entered the second set of iterations. Here, the university team introduced annotation tasks for the industry team. A customised Prodigy by spaCy \citep{prodigy} NER annotation interface was used for annotating; more information can be found in Appendix~\ref{sec:appendix:anno_gui}.

After the industry team completed these tasks, their feedback was discussed to further refine the guidelines. This feedback included analysis of challenging cases and observations on whether additional aspects should be incorporated. Once the revised guidelines were in place, the domain-expert researcher who had led their development (annotator \ann{ForestGreen}{A2}) annotated every advert, including the 113 we later used to measure inter-annotator agreement (Section~\ref{sec:iaa}).


\subsection{Span Labels}

As a result of this process, we reached a consensus on three main span labels (these may also be referred to as `classes' or `tags') for our study. These tags were designed to classify text abstractly, capturing a broader portion of the content with a small number of labels. Specifically, we identified three tags:

\begin{itemize}[leftmargin=*,itemsep=0pt,parsep=0.2pt]
    \item \underline{Condition Tag:} refers to the current condition of the vehicle. Instances of this may include engine noise, the presence of scratches, and tyre condition. The span would typically include a noun and an attribute (e.g., ``tyre tread is low''). 
    
    \item \underline{Historic Tag:} refers to an event that happened to the car before the posting of the ad. Instances of these include past accidents, number of previous owners, component change and servicing (e.g., ``new cambelt fitted last August''). 

    \item \underline{Sales Options Tag:} refers to any tangible item or service offered during the sale --- apart from the car. These could refer to part exchange offers, delivery, warranties and test drives. 
    
    
\end{itemize}


\paragraph{Why three labels?} The labels separate three readings that the same mention can carry: a state of the car now (\textit{Condition}), an event in its past (\textit{Historic}), or something offered with the sale (\textit{Sales Options}). Table~\ref{tab:families} (Appendix~\ref{sec:appendix:families}) shows four frequent mention types under all three labels. The very string ``12 months MOT'' is annotated six times as \textit{Condition} (the certificate the car holds) and six times as \textit{Sales Options} (a promise at the point of sale). The distinction is contextual rather than lexical, which makes the task hard even for people: of the 586 spans on which the two expert annotators agreed on boundaries, they disagreed on the label for seven, six of them MOT or service mentions (e.g., ``Recent service'', read as \textit{Historic} by one and \textit{Condition} by the other). Collapsing the three labels would lose exactly what downstream users need to know: whether an MOT is valid now, was passed recently, or is being offered.

\noindent

In Appendix~\ref{sec:appendix:labels}, we disclose the annotation guidelines, which provide the definitions of each label, detected sub-types and span labelling rules. Furthermore, additional span types encountered during our annotation phase were excluded from the study for reasons detailed in the appendix (see Appendix~\ref{sec:appendix:not_tracked} for documentation).

\label{sec:prel}
\subsection{Dataset Origin}
We sourced the data from our industry partner's proprietary database of current car advertisements. Half of the texts of these advertisements are written by retailers, covering \textit{traders}, car dealerships with a large available stock of vehicles and high turnover and the other half by \textit{individuals} typically infrequently listing a single vehicle.

In addition, to ensure a representative sample for the trader dataset, we sorted traders by the number of their listings and selected those with the highest number of listings, as they typically follow a template for their sale ads. This approach ensured a stratified sample of car-trader listings, capturing both those that follow a standardised template and those that write listings freely. For the private dataset, we stratified the listings by UK region (e.g., ``North West'') and sampled a proportional number of car listings from each region. This method ensured that different regional variations of the English language are represented in our dataset.

An example of a short advert is provided within the annotation interface, as shown in Appendix~\ref{sec:appendix:anno_gui}, Figure~\ref{fig:prodigy}.



\subsection{Dataset Overview}
\label{sec:data_attrs}
In our dataset, a total of 605 advertisements were analysed, amounting to 104,382 tokens (spaCy's rule-based English tokenisation, which we use for annotation, the CRF and evaluation throughout). Notably, 30 of these advertisements (approximately 5\%) contained no annotatable spans. These ads typically consisted solely of sales pitches, car standard/optional features, or copied the header (e.g., ``2.0-litre engine BMW''). In Table~\ref{tab:stats}, we present the total number of labels and the spaCy tokens they cover for each label in our dataset.

\begin{table}[t]
\centering

\resizebox{\columnwidth}{!}{%
\begin{tabular}{llll}
\hline
Number of & CONDITION & HISTORIC & \begin{tabular}[c]{@{}l@{}}SALES\\ OPTIONS\end{tabular} \\ \hline
\# Labels & 594       & 810      & 2,174                                                  \\
\# Tokens (spaCy) & 2,466       & 3,985    & 6,216                                                  \\
\# Sub-word tokens (DeBERTaV3) & 2,957       & 4,559    & 6,875                                                  \\ \hline
\end{tabular}%
}
\caption{Distribution of labels in our dataset, with the spaCy word tokens they cover and the corresponding sub-word tokens of the \texttt{deberta-v3-base} vocabulary.}
\label{tab:stats}
\end{table}


Compared to other sequence labelling datasets from the literature, our dataset would be most similar to ``noisy'' datasets such as WNUT16 \citep{strauss-etal-2016-results} and Twitter NER \citep{ritter-etal-2011-named}. The similarities lie in the use of colloquial language, typos, non-standard formatting, abbreviations, and some instances of emoticons. However, the unique aspect of our dataset is that it is designed for users to submit sales pitches for their cars. Consequently, distinct features include the use of telegraphic-style language, non-standard grammar (e.g., missing connecting words), the omission of contextual information and capitalisation to capture the buyer's attention. Additionally, we observed extensive use of persuasive language (e.g., positive adjectives), direct calls to action (e.g., imperative sentences encouraging viewers to call or not to miss out on the opportunity), and detailed lists of features, attributes or services.

Moreover, we observed a clear distinction between texts authored by traders and those written by individuals. Traders typically adopt a customisable, template-based approach to ensure professionalism and efficiency, whereas individuals’ posts vary widely in style and tone. Furthermore, traders often enumerate all \textit{Sales Options} they offer using bullet points or lists, while individuals generally do not provide such options.

To gain deeper insight into the linguistic characteristics associated with each class, we conducted a comparative analysis of part-of-speech (PoS) tag distributions across the three labels using spaCy \citep{Honnibal2020}. The \textit{Sales Options} label is characterised by a notably higher proportion of singular nouns (37.7\%) and proper nouns (25.7\%), reflecting a strong emphasis on the financial products offered by retailers. In contrast, the \textit{Historic} label demonstrates a more balanced distribution, showing substantial usage of singular nouns (23.9\%), adjectives (14.9\%), proper nouns (14.5\%), and cardinal numbers (13.4\%), which indicates a descriptive narrative style focusing on historical events, proper names, dates, and the vehicle's number of owners. 

The \textit{Condition} label similarly exhibits balanced noun usage (23.7\%) but places greater emphasis on cardinal numbers (14.1\%) and prepositions or subordinating conjunctions (10.2\%). These patterns suggest a descriptive and relational linguistic structure oriented towards vehicle conditions, states, current mileage, and MOT expiry dates. Overall, these distinctions highlight slightly differing linguistic profiles, with \textit{Sales Options} being more noun-focused, \textit{Historic} event- and description-oriented, and \textit{Condition} emphasising descriptive and relational language. Table~\ref{tab:pos_percentages} in Appendix~\ref{sec:appendix:pos} summarises the most common PoS tags per label.

\section{Experiments}
\label{sec:exp}
\subsection{Evaluation Metric}
\label{sec:exp:evaluation}
For evaluation, we used an adjusted $F_1$-score for NER. We report the partial match based on \citet{segura-bedmar-etal-2013-semeval} who take into consideration various categories of errors \citep{chinchor-sundheim-1993-muc}. Specifically, they report:
\begin{enumerate}
\itemsep0em
    \item $COR$rect, when the spans between the gold standard and the prediction match entirely
    \item $INC$orrect, when the spans match, but the label is different
    \item $PAR$tial, when the spans overlap and have the same label
    \item $MIS$sing, when a gold annotation span is not predicted
    \item $SPU$rious, when a model predicts spans which do not exist in the gold standard.
\end{enumerate}
Following this, we define the possible and actual matches as follows:

{\small
\begin{displaymath}
\begin{aligned}
POS\text{sible} &= COR + INC + PAR + MIS \\
&= TP + FN \\
ACT\text{ual} &= COR + INC + PAR + SPU \\
&= TP + FP
\end{aligned}
\end{displaymath}
}


The precision and recall can be characterised as:
{\small
\begin{displaymath}
\begin{aligned}
\text{Precision} &= \frac{COR + 0.5 \times PAR}{ACT} \\
&= \frac{TP}{TP+FP}
\end{aligned}
\end{displaymath}
}
{\small
\begin{displaymath}
\begin{aligned}
\text{Recall} &= \frac{COR + 0.5 \times PAR}{POS} \\
&= \frac{TP}{TP+FN}
\end{aligned}
\end{displaymath}
}
\noindent
We utilise the $F_1$-score (the harmonic mean between precision and recall) for IAA by applying it with one annotator serving as the gold standard against the other. Cohen's Kappa is widely recognised as the preferred metric for IAA, having been developed to address the shortcomings of percentage agreement by accounting for chance agreement \citep{McHugh2012}. However, research has pointed out its unsuitability for NER, attributing this to the lack of true negatives \citep{brandsen-etal-2020-creating, richie-etal-2022-inter}.

\subsection{Inter-Annotator Agreement}
\label{sec:iaa}
The IAA study involved three annotators, all authors of this paper: a domain-expert data scientist from the industry team (\ann{RoyalBlue}{A1}); a university-based researcher who led the creation of the annotation guidelines (\ann{ForestGreen}{A2}); and a university-based, non-automotive domain-expert researcher who joined after the guidelines were finalised (\ann{BrickRed}{A3}). 


The IAA was calculated on 113 adverts annotated by all three annotators, representing approximately 19\% of the total corpus (605 ads). The mean pairwise partial-match $F_1$-score was 88\% for the \ann{RoyalBlue}{A1}--\ann{ForestGreen}{A2} pair, 73\% for \ann{RoyalBlue}{A1}--\ann{BrickRed}{A3}, and 73\% for \ann{ForestGreen}{A2}--\ann{BrickRed}{A3}, achieving an overall score of $\sim$78\%. A detailed breakdown of agreement by category averaged across all the annotators is presented in Table~\ref{tab:res_labels}. More details on the exact and partial matching scores per label are provided in Table~\ref{tab:iaa-scores} (Appendix~\ref{sec:app:iaa}).

\begin{table}[h]
\centering
\resizebox{\columnwidth}{!}{%
\begin{tabular}{lccccc}
\toprule
\textbf{Label} & \textbf{COR} & \textbf{INC} & \textbf{PAR} & \textbf{MIS} & \textbf{SPU} \\
\midrule
CONDITION    & 72\% & 7\% & 12\% & 5\% & 5\% \\
HISTORIC     & 63\% & 5\% & 20\% & 6\% & 6\% \\
SALES\_      & \multirow{2}{*}{60\%} & \multirow{2}{*}{2\%} & \multirow{2}{*}{22\%} & \multirow{2}{*}{8\%} & \multirow{2}{*}{8\%} \\
OPTIONS      &  &  &  &  &  \\
\bottomrule
\end{tabular}
}
\caption{Pairwise inter-annotator agreement averaged across all three annotators for each label. Scores are computed over all ordered annotator pairs and report the proportion of correct (cor), incorrect (inc), partial (par), missing (mis), and spurious (spu) span matches following the evaluation scheme in Section~\ref{sec:exp:evaluation}.}
\label{tab:res_labels}
\vspace{-1.5em}
\end{table}



\subsection{Models}
For our experiments, we compared in total seven models. Following, we have grouped them by type.

\subsubsection{Encoder Transformer Models}
We tested two types of encoder architecture transformers. One was a BERT \citep{devlin-etal-2019-bert} model, the other was a DeBERTaV3 \citep{he2021debertav3} model. We trained them according to the token-classification methodology\footnote{\href{https://github.com/huggingface/transformers/tree/main/examples/pytorch/token-classification}{token-classification}} from HuggingFace \citep{wolf2020huggingfacestransformersstateoftheartnatural}. Listing~\ref{lst:hyperparameters} in Appendix~\ref{sec:appendix:hyper-params} details the hyper-parameters used in our experiments for these two models.


\subsubsection{Decoder Transformer Models}
We experimented with two open-source and two closed-source decoder-based transformer large language models. Specifically, these were: GPT-4o \citep{openai2024gpt4}, Gemini Flash \citep{geminiteam2024gemini}, Llama \citep{grattafiori2024llama3herdmodels} and Qwen \citep{yang2024qwen2technicalreport}. 
Their precise versions are in Listing~\ref{lst:llm_versions} (Appendix~\ref{sec:appendix:hyper-params}).
The full prompt comprises three parts, shown in Figures~\ref{fig:prompt_si}--\ref{fig:prompt_sample} in the order in which they appear: first, the system instruction (Figure~\ref{fig:prompt_si}); second, the task and label definition (Figure~\ref{fig:prompt_guidelines}); and finally, the in-context examples followed by the advertisement to be annotated (Figure~\ref{fig:prompt_sample}).

\begin{figure}
    \centering
    \begin{tcolorbox}[colback=gray!5!white,colframe=gray!75!black,title=System Instruction \hfill Part 1/3]
You are a Natural Language Processing assistant specializing in car-related named entity recognition. In the following conversation, perform the tasks exactly as requested, without providing explanations or reasoning.
\end{tcolorbox}
    \caption{Prompt part 1/3: the system instruction used for the LLMs.}
    \label{fig:prompt_si}
\end{figure}

The system-instruction prompt (Figure~\ref{fig:prompt_si}) constrained the model’s outputs to minimise verbosity, suppress explicit reasoning and justification of choices, and focus exclusively on task execution. It also included a role-playing instruction assigning the model the role of an NLP assistant, which prior work has shown to be effective \cite{kong-etal-2024-better}.

\begin{figure}
    \centering
   \begin{tcolorbox}[colback=white,colframe=black!75!black,title=Label Definition \hfill Part 2/3]
\textbf{User:} The task is to label <label> entities in the given text. <label> is characterised as `<label\_definition>'. Your task is to detect <label> span words by inserting `@@' at the beginning of the span and `\#\#' at the end, and generating the remaining text verbatim, whitespaces and newlines included for parsing purposes. \\
\textbf{Assistant:} Sure, I can help with that. Please provide the text you want me to label for <label> entities.
\end{tcolorbox}
    \caption{Prompt part 2/3: the prompt that introduces the task and the label's definition (label\_definition). The label definitions are in Appendix~\ref{sec:appendix:labels}.}
    \label{fig:prompt_guidelines}
\end{figure}

For prediction with the LLMs, we followed the NER setting as proposed by \citet{wang2023gptnernamedentityrecognition}.
We employed random 100 samples from the training data in an in-context learning (ICL) scenario. Specifically, for each label, we empirically added the label's definition at the beginning of the prompt; one can see our prompt template in Figure~\ref{fig:prompt_guidelines}.

\begin{figure}[h]
    \centering
    \begin{tcolorbox}[colback=white,colframe=black!75!black,title=ICL Prompt \& Inference \hfill Part 3/3]
\textbf{User:} We offer 6 month warranty. \\
\textbf{Assistant:} We offer @@6 month warranty\#\#. \\
\textbf{User:} We provide Full Service history.\\
\textbf{Assistant:}  We provide @@Full Service history\#\#.\\
\textbf{User:} 12 month warranty is included in the sale. 
\end{tcolorbox}
    \caption{Prompt part 3/3: the in-context examples and the advertisement to be annotated. The figure shows only two ICL examples for illustration. These would typically be longer—an entire car advert, which might comprise one to several sentences—rather than a single sentence. The last \textbf{User} utterance points to a sentence (i.e., advertisement) that we expect the LLM to annotate with the special characters ``@@'' and ``\#\#'' to obtain the token spans that correspond to the label of interest.}
    \label{fig:prompt_sample}
\end{figure}

Then we used a chat alternating history to demonstrate how a text should be labelled. At the end of the prompt, we included a sample from the test dataset to collect the generated output. The output would replicate the input text, but the spans would be marked by prefixing with `@@' and suffixing with `\#\#' to segment them from the rest of the text. An example is shown in Figure~\ref{fig:prompt_sample}. In cases where the text was not copied verbatim, we tackled this issue using an alignment library\footnote{\href{https://pypi.org/project/pytextspan/}{pytextspan}}.


\subsubsection{Conditional Random Fields}
Lastly, we employed CRF models \citep{crf} with different feature configurations, utilising a range of lexical, morphological, and part-of-speech tags (PoS) obtained through the spaCy NLP toolkit \citep{Honnibal2020}. The features for the tokens were encoded with the following attributes extracted from spaCy: 
\begin{multicols}{2}\raggedcolumns
\begin{itemize}[leftmargin=*,itemsep=0pt,topsep=0pt] 
\item lowercased form 
\item suffixes (last two and last three characters) 
\item whether it is capitalised or title-cased 
\item whether it is a digit 
\item token shape (pattern of character types) 
\item whether it is alphabetic 
\item whether it is a stop word
\item both coarse- and fine-grained PoS tags.
\end{itemize}
\end{multicols}

Furthermore, we incorporated contextual features from the immediately preceding and following tokens (i.e., ±1, ±2, ±3): their lowercased forms, coarse-grained PoS tags, and digit indicators. To fine-tune the CRF model’s hyperparameters—specifically, the L1 and L2 regularisation coefficients—we conducted a randomised 100-iteration hyperparameter search on the validation set.

On this feature set we implemented four configurations: 
\begin{enumerate} [itemsep=0pt, topsep=0pt]
\item CRF: includes all of the features, with a ±1 token context window. 
\item CRF PoS: includes only spaCy’s PoS tags. 
\item CRF w/o PoS: includes all features (as in 1.) except spaCy’s PoS tags. 
\item CRF ±2/±3: identical to 1. but with broader context windows (±2 and ±3 tokens). 
\end{enumerate}

We included the PoS-only setting (2) and our first setting without the PoS tags (3) to evaluate the influence of PoS tags since our data overview analysis, in Section~\ref{sec:data_attrs}, indicated slightly different PoS distributions across labels.


\subsubsection{Computational Resources}
All training and inference for the open-source models were performed on NVIDIA H100 Hopper graphics cards, except for the CRF implementations, which, due to their lower computation requirements, were run on an Apple M1 Max chip.\footnote{Hardware specifications: \url{https://www.nvidia.com/en-gb/data-center/h100/} and \url{https://support.apple.com/en-gb/111902}.} 

\subsection{Results}
For our experiments, we measured performance across three folds of the dataset, where the test sets were mutually exclusive. Each fold used 70\% of the instances for training, with 15\% each for validation and testing. The LLMs would not require the validation dataset, since we did not fine-tune them. In some cases, the LLMs' generations were incorrect and could not be recovered, even with our alignment library; this occurred rarely, less than 1\% in each fold. The overall results averaged across the three folds are presented in Table~\ref{tab:res}. Table~\ref{tab:res_per_label} breaks the $F_1$-score down per label for the four best-performing models, and Section~\ref{sec:error_analysis} decomposes the matches of those four models into the five categories of Section~\ref{sec:exp:evaluation}.



One can view in the table that the fine-tuned DeBERTaV3 model, trained on the full training split, obtains the highest $F_1$-score. Looking more closely at the table, the precision is much higher than the recall for the decoder models. This may be because we utilised only 100 random ICL examples per fold. We took this approach to maintain fairness in comparing the models and reduce computation time and cost.

We noticed that GPT-4o did not annotate a particular span(s) at various times (i.e., low recall). LLMs may be susceptible to errors in previous generations or context, affecting subsequent text generations \citep{10.1145/3744238}. Hence, we hypothesise that multiple LLM agents with different ICL examples could yield better results after aggregating the outputs. Another technique could be to select more representative ICL samples per advertisement to label.

The CRF approach outperformed every LLM and BERT, highlighting its effectiveness in low-resource, domain-specific tasks. Analysis of the trained CRF model showed that its most influential features were lexical in nature. For instance, terms such as ``FSH'' and ``warranties'' strongly indicated the \textit{Sales Options} label. This observation aligns with the results presented in Table~\ref{tab:res}, where the CRF implementation employing default features performed best, while the model relying solely on PoS features performed poorly. The ten most influential features per label identified by the CRF implementation are listed in Appendix~\ref{sec:appendix:crf_top}. 

\begin{table}[ht]
\small
\centering

\begin{tabular}{cccc}
\hline
\textbf{Model}                                              & \textbf{Precision}                & \textbf{Recall}                   & \textbf{F1-score}                 \\ \hline
BERT-base                                                   & 52.6 $\pm$1.7 & 63.5 $\pm$2.4 & 57.5 $\pm$1.2 \\
\begin{tabular}[c]{@{}c@{}}DeBERTaV3-\\ base\end{tabular}   & 63.6 $\pm$1.2 & \textbf{72.4 $\pm$2.0} & \textbf{67.7 $\pm$0.8} \\
\hline
GPT-4o                                                      & 73.8 $\pm$1.6 & 54.4 $\pm$1.3 & 62.6 $\pm$0.3 \\
\begin{tabular}[c]{@{}c@{}}Gemini 1.5 \\ Flash\end{tabular} & 72.4 $\pm$2.1 & 49.9 $\pm$3.2 & 59.0 $\pm$2.0 \\
\begin{tabular}[c]{@{}c@{}}Llama 3.1 \\ 70B it\end{tabular} & 75.9 $\pm$1.1 & 44.5 $\pm$2.3 & 56.1 $\pm$2.1 \\
\begin{tabular}[c]{@{}c@{}}Qwen 2.5 \\ 72B it\end{tabular}  & \textbf{76.8 $\pm$3.2} & 34.8 $\pm$1.1 & 47.9 $\pm$1.6 \\ \hline
CRF                                                   & 72.5 $\pm$1.8 & 60.8 $\pm$1.5 & 66.1 $\pm$1.3 \\

CRF PoS                                                  & 25.4 $\pm$8.6 & 1.1 $\pm$0.1 & 2.1 $\pm$0.2 \\

CRF w/o PoS                                                  & 71.9 $\pm$0.9 & 58.5 $\pm$2.5 & 64.4 $\pm$1.7 \\

CRF 2n/3n                                                  & 71.2 $\pm$0.8 & 60.1 $\pm$0.9 & 65.2 $\pm$0.8 \\

\hline
\end{tabular}
\caption{Mean and population standard deviation over the three folds of precision, recall and $F_1$-score (\%) for each model; ``it'' stands for instruction-tuned; bold marks the highest value in each column.}
\label{tab:res}
\end{table}

\begin{table}[h]
\centering
\small
\begin{tabular}{lccc}
\toprule
\multirow{2}{*}{\textbf{Model}} & \multicolumn{3}{c}{\textbf{F1-score (mean$\pm$std)}} \\
\cmidrule(lr){2-4}
 & \textbf{Historic} & \textbf{Sales Option} & \textbf{Condition} \\
\midrule
\begin{tabular}[c]{@{}c@{}}DeBERTaV3-\\ base\end{tabular} & 58.1$\pm$2.3 & \textbf{74.0$\pm$2.7} & \textbf{57.8$\pm$5.7} \\
\midrule
GPT-4o & 53.0$\pm$2.2 & 68.4$\pm$1.9 & 52.2$\pm$5.2 \\
\begin{tabular}[c]{@{}c@{}}Gemini 1.5 \\ Flash\end{tabular}  & 51.6$\pm$2.4 & 66.7$\pm$4.1 & 38.5$\pm$5.2 \\
\midrule
CRF & \textbf{59.5$\pm$3.8} & 73.2$\pm$2.8 & 46.5$\pm$4.4 \\
\bottomrule
\end{tabular}
\caption{$F_1$-score performance across different labels for the top four performing models. Values are shown as percentages. The highest scores in each column are bolded.}
\label{tab:res_per_label}
\end{table}

\subsection{Error Analysis}
\label{sec:error_analysis}
Table~\ref{tab:error_types} decomposes the matches of the four best-performing models into the five categories of Section~\ref{sec:exp:evaluation}; Appendix~\ref{sec:appendix:error_types} gives the same breakdown per label. It resolves an apparent tension in Table~\ref{tab:res}: the LLMs enclose spans more tightly, yet score lower. Their precision is the highest because a span they do emit is more often exact and less often spurious---partial matches make up 14.6\% of GPT-4o's matches against 17.2\% for the CRF and 19.8\% for DeBERTaV3---but they leave far more gold spans unannotated (31.2\% of GPT-4o's and 35.9\% of Gemini's matches are missing spans, against 23.2\% for the CRF and 7.7\% for DeBERTaV3). Since a partial match still earns half credit whereas a miss earns none, the models that find more spans with looser boundaries come out ahead on $F_1$. The boundary errors also differ in kind: when the CRF or GPT-4o produces a partial match, the predicted span is more often too wide than too narrow (58\% against 33\% of the CRF's partial matches, 58\% against 38\% of GPT-4o's), whereas DeBERTaV3's partial matches are more often too narrow (53\%). Per label (Table~\ref{tab:error_types_per_label}), \textit{Condition} is the hardest for every model, with the highest shares of label confusions and of misses: GPT-4o and the CRF each fail to annotate about 40\% of the gold \textit{Condition} spans, Gemini about 54\%, and DeBERTaV3 about 12\%. The contextual mention types of Table~\ref{tab:families} are where label confusions concentrate: on gold spans mentioning an MOT, the four models assign the wrong label to 15.2\% (DeBERTaV3), 19.3\% (GPT-4o), 27.1\% (Gemini) and 19.1\% (CRF) of the spans they find, between four and seven times their rates on spans outside these types (3.0--5.0\%); service mentions show the same pattern at 10.9--14.5\%. The models also split where the experts split: on ``12 month MOT'', gold \textit{Sales Options}, all four predicted \textit{Condition}, the reading the other expert annotator had chosen. Finally, although not the top scorer, the CRF model may be preferable to transformer-based approaches due to its greater algorithmic accountability and transparency \citep{Diakopoulos}.

\begin{table}[h]
\centering
\small
\resizebox{\columnwidth}{!}{%
\begin{tabular}{lccccc}
\toprule
\textbf{Model} & \textbf{COR} & \textbf{INC} & \textbf{PAR} & \textbf{MIS} & \textbf{SPU} \\
\midrule
DeBERTaV3-base & 48.7 & 4.9 & 19.8 & 7.7 & 18.8 \\
GPT-4o & 43.3 & 3.9 & 14.6 & 31.2 & 7.0 \\
Gemini 1.5 Flash & 40.6 & 4.6 & 11.2 & 35.9 & 7.6 \\
CRF & 47.0 & 4.1 & 17.2 & 23.2 & 8.4 \\
\bottomrule
\end{tabular}}
\caption{Match types per model as percentages of its matches pooled over the three test folds, following Table~\ref{tab:res_labels} (COR correct, INC incorrect label, PAR partial overlap, MIS missing, SPU spurious); Appendix~\ref{sec:appendix:error_types} gives the breakdown per label.}
\label{tab:error_types}
\end{table}

\paragraph{Qualitative Examples} The CRF and GPT-4o performed worse than human annotators at delineating contextual cues. For instance, in the example ``This is my lovely car that I have used for \underline{14000 miles}'' the CRF mistakenly treated the miles as the car's mileage rather than the distance the last owner had driven (a label confusion, \textit{Condition} for \textit{Historic}). Whereas, in the nuanced example of ``we will add new carpets upon purchase'', the CRF correctly predicted the statement as \textit{Sales Options} since it is added value at the point of sale, but GPT-4o labelled the part as \textit{Condition}, which is a less favourable label since it is not a current statement about the car but rather an added sales feature (again a label confusion). Moreover, in an example where the owner had written the mileage stamps for when their car was serviced, the CRF falsely labelled them as \textit{Condition} (a label confusion), and they were missed by GPT-4o (a missing span), which should have been labelled as \textit{Historic}. This suggests that both approaches have merits and downsides; perhaps an ensemble of the two could yield more favourable results. To add to that claim, it is evident in Table~\ref{tab:res_per_label} that the CRF demonstrates a higher average $F_1$-score than GPT-4o for predicting the \textit{Historic} and \textit{Sales Options} labels, whereas GPT-4o scored higher on the \textit{Condition} label.

\section{Discussion}
\label{sec:discussion}
We present the first annotation schema for car advertisements, meticulously developed to meet genuine industry demands in a real industry setting. We established three span labels and corresponding guidelines after two rounds of collaborative iterations between university researchers and industry partners. We then evaluated inter-annotator agreement, achieving a commendable 88\% $F_1$-score between the domain-expert annotators and 78\% overall.


One theme that emerged from Table~\ref{tab:res_labels} was that there were slightly more partial matches for the \textit{Sales Options} and \textit{Historic} labels (i.e., only segments of a span are matched against the ground truth). Although we made efforts to mitigate this issue by editing the annotation guidelines, we attribute it to differences in interpretation when labelling. \textit{Historic} labels often refer to past events, leading annotators to include additional, possibly extraneous, contextual information. \textit{Sales Options} labels, owing to the liberal use of adjectives to describe services a dealer offers, create ambiguity about which aspects are sufficiently factual or necessary for inclusion. 

\paragraph{Potential Use Cases} The three labels allow us to conduct multiple analyses. Firstly, the \textit{Sales Options} enable us to examine, across temporal and geographical dimensions, which sales options/features are offered, such as warranties, financing, and part-exchanges, which may vary significantly depending on economic conditions and regional preferences. This raises questions about whether a robust economy alters these trends and what consumers expect from their dealings with car dealerships.

These labels can help us better understand which information writers typically include in advertisements and, in marketing terms, what makes a listing more appealing. For example, they can show whether the \textit{Condition} of a vehicle (e.g., the presence of new tyres or the extent of scratches) plays a critical role in shaping consumer decisions, and whether \textit{Historic} labels, like the number of previous owners or a vehicle's maintenance history, are valued across different regions.

This study serves as a cornerstone for downstream tasks, since it delineates the nuanced readings of Table~\ref{tab:families} and would make further sub-categorisations simpler. We identified several sub-types of mentions emerging from our data. A non-exhaustive list of these sub-types is documented in Appendix~\ref{sec:appendix:labels}. Looking ahead, we aim to explore \textbf{entity typing}, where the type of each span is generated as free text rather than selected from a closed ontology \citep{choi-etal-2018-ultra,yuan-etal-2022-generative-entity}.

\paragraph{Wider Applicability} The application of such span labelling extends to safety and accident analysis, where patterns categorised by location and severity provide essential insights for police, insurance companies, and urban planners considering road expansions.

One can use our work to develop tools to educate car owners about necessary or upcoming maintenance, potentially revolutionising consumer knowledge and empowering car owners. Moreover, it could be applied to quantify and highlight instances of under-reporting and suggest policies to enhance consumer awareness. Such comprehensive use of data ensures a more transparent marketplace and significantly enhances consumer safety and satisfaction.

We believe the text-mining applications enabled by the proposed span-labelling schema can benefit various stakeholders in the vehicle trading industry. The schema can offer more accurate evaluations of vehicle features, auto-complete suggestions when writing an advert, and evaluation of advertisement quality.

\section{Conclusion}
\label{sec:concl}
Our effort represents the first-ever English-language annotation schema and dataset for information extraction in the automotive advertisement domain. We collaboratively developed the span annotation schema with involvement from industry and academia. The three labels chosen allow us to annotate meaningful spans that alter the car's value or affect the sale proposition. We achieved an encouraging IAA score and observed that a fine-tuned DeBERTaV3 encoder performed best, closely followed by the CRF, with Qwen leading in precision. The CRF demonstrates its adaptability on this low-resource domain-specific task, surpassing all of the LLMs. Lastly, after analysing the models, we elaborate on different uses of this schema from insurance companies to empowering car owners.

\section*{Limitations}
As a foundational case study introducing a novel domain schema, this work has certain practical limitations that pave the way for future research. 

Firstly, the manually annotated dataset of 605 advertisements was sufficient to establish the \textbf{Auto-AdvER} schema, reach consensus, and benchmark our baseline models, though a larger corpus would capture more of the long tail of linguistic variation in the UK automotive market. Because the advertisements are written in British English, they also reflect UK-specific terminology and conventions that may differ slightly from those in other English-speaking regions.

Secondly, reflecting the genuine composition of retail vehicle advertisements, there is a natural label imbalance within our dataset. Because sellers typically enumerate numerous services and perks to attract buyers, \textit{Sales Options} naturally occur more frequently than \textit{Condition} or \textit{Historic} mentions. Although this represents a realistic data distribution, it poses a standard learning challenge for supervised models, which could be mitigated in future iterations through targeted sampling.


Finally, regarding our experimental setup, the use of closed-source LLMs complicates direct comparisons with open-source models due to their undisclosed training data. This lack of transparency can lead to potential data contamination issues, obscuring true performance differences and making baseline evaluations against open-source models incomplete.

\section*{Ethics Statement}

In conducting this research, our primary motivation is to contribute to the greater good by enabling more informed consumers and enhancing the capabilities of individuals selling cars. We believe that a robust annotation schema for car advertisements will empower consumers with clearer, more reliable information and provide sellers with a tool to better present their products. Additionally, we acknowledge the ethical considerations associated with our work. We utilised LLMs. While these and similar models offer significant advancements, it is important to remain mindful of their substantial energy consumption and the environmental impact associated with their extensive use. Therefore, we advocate a balanced approach, ensuring that the benefits of such technologies are weighed against their ecological footprint.


\section*{Acknowledgements}
We thank our industry partner, Autotrader UK, for providing the car advertisement data, for sharing the domain expertise that shaped the \textbf{Auto-AdvER} schema and dataset, and for the funding that supported this research.

\bibliography{output}

\appendix
\section{Annotation GUI}
\label{sec:appendix:anno_gui}
\begin{figure}[ht]
    \centering
    \includegraphics[width=\columnwidth]{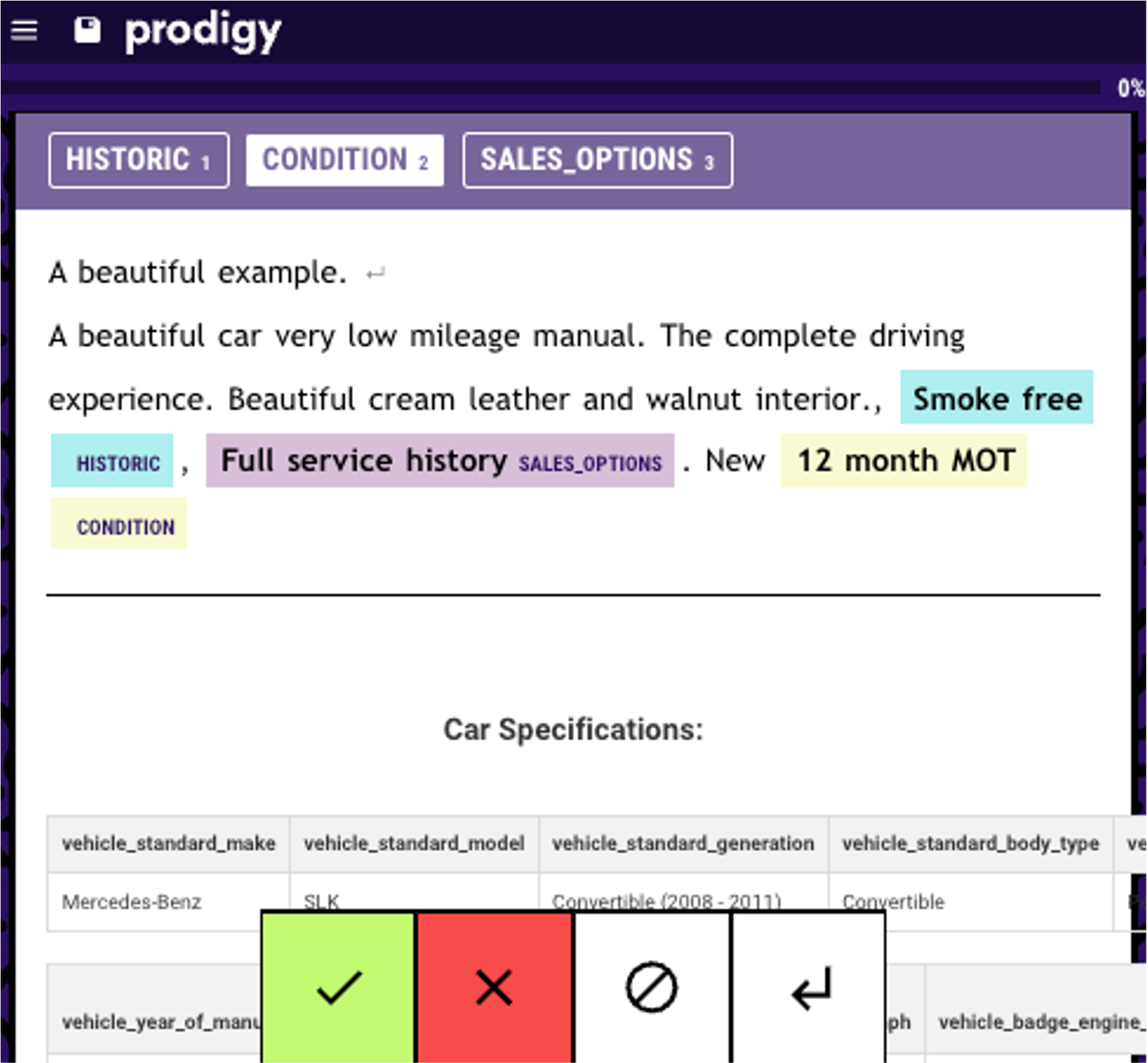}
    \caption{An example of the annotation interface, illustrating a customised NER annotation task implemented using Prodigy, displayed within a reduced-size window.}
    \label{fig:prodigy}
\end{figure}

For annotating we used Prodigy by spaCy \citep{prodigy}. We implemented a customised sequence labelling version, which let our annotators choose amongst the three labels. Also, below the annotation text, the interface will show the vehicle's specifications that are being annotated for context. Figure \ref{fig:prodigy} shows an example of an annotated text.

\section{Label's Definitions \& Identified sub-entities}
\label{sec:appendix:labels}

Below, are the annotation guidelines used for our study, along with the sub-entities detected.

\subsection*{Condition - Current state}
The \textit{Condition} category encompasses any valuable car-related information that is current as of the date the advertisement was posted online.
\begin{tcolorbox}[colback=gray!5!white, colframe=gray!75!black, title=Definition]

The \textit{CONDITION} label identifies any word or span of words in car advertisements that indicate the present material or documented condition of the vehicle, both positive and negative. This includes descriptions that reflect the car's current operational performance, mechanical state, aesthetic appearance, maintenance status, and legal classifications affecting its value or usability
\end{tcolorbox}
These include:
\begin{itemize}
    \item Tyre condition (i.e. trim level or mention of good condition)
    \item Interior condition (e.g., ``leather like new'')
    \item Exterior condition (e.g., ``presence of scratches'')
    \item Mechanical condition (e.g., ``engine noise'', ``gear not locking properly'')
    \item MOT\footnote{This refers to the Ministry of Transport, which, in a broader context, represents any legal authority responsible for ensuring vehicle safety by enforcing compliance with minimum standards.} due date or recent done MOT mentioned
    \item Service due date or recent done Service mentioned
    \item Current mileage (e.g., ``12k miles'')
    \item Cat S, C, D or N\footnote{This refers to vehicle categories (such as Cat S, C, D, or N), which, in a broader context, denote classifications used by insurance companies and authorities in the UK to indicate vehicles that have sustained damage and have subsequently been repaired or written off.}
    \item Registered, pre-registered, non-registered, demonstrator\footnote{This refers to vehicle registration statuses (such as registered, pre-registered, non-registered, or demonstrator), which, in a broader context, indicate whether a vehicle has been officially recorded with authorities, previously assigned registration by a dealership, remains unregistered, or has been used primarily for demonstration purposes.}
\end{itemize}

\paragraph{Annotation Span Length \& Methodology:} \begin{enumerate}
\item A condition span is typically characterised by the \underline{noun} of part of a vehicle (e.g., ``tyre thread'') and an \underline{attribute} that links to (e.g., ``low''). You would annotate the whole phrase (e.g., ``tyre thread is low''). And omit anything else.
\item We would disregard statements that are opinion based, such as ``low mileage'', since it is a statement denoting an opinion that the car is of low mileage of its age. Instead ``80 000 mileage'' would be labelled, since it's presented as a fact. In a similar fashion ``amazing condition'' does not add enough valuable information to the ad, being characterised of opinion, whereas ``clean paintwork'' is specific and adds value to the ad, hence, we would label the second one but not the first.
\item Remember to label span words that are referring to the state of the vehicle when the ad was posted e.g., ``6 month MOT'', ``80 000 miles'', references to doing extra miles after the ad was posted can be safely ignored.
\end{enumerate}

\subsection*{Historic - Events in the past}
The \textit{Historic} attribute pertains to events that have occurred prior setting the car advertisement. This category focuses on the events that happened.


\begin{tcolorbox}[colback=gray!5!white, colframe=gray!75!black, title=Definition]
The \textit{HISTORIC} label identifies any word or span of words that presents events that have happened to the car in the past and affected its condition mechanically condition or bodywork.

\end{tcolorbox}

These include:
\begin{itemize}
    \item Mention of past accidents (e.g., ``3 years back I scratched the front side'')
    \item Number of owners
        \begin{itemize}
        \item  Solely without mentioning ``from new''. E.g., ``3 owners \sout{from new}''
        \end{itemize}
    \item Past type of usage 
    \begin{itemize}
        \item Any mention of past usage that could affect its condition, either positive or negative. E.g., ``smoke free'', ``motorway mileage'', ``stored in locked garage''
    \end{itemize}
  
    \item Component change (e.g., ``Cambelt changed a year ago'', ``New brake callipers'', ``new tyres, brake discs'') and tuning (e.g., ``re-mapping'')
    \item MOT advisories or no advisories mentioned
    \item HPI\footnote{This refers to Hire Purchase Investigation, which, in a broader context, denotes a vehicle history check commonly used in the UK to verify whether a vehicle has outstanding finance, has been reported stolen, previously written off, or has other hidden issues.} \underline{is} clear (as opposed to only providing HPI report as a Sales Option, see further)
    \item Service history done in the past, including stamps (e.g., 8 stamps), dates (e.g., ``02/2023, 01/2024'') or on which mileages (e.g., ``@ 30k, @40k, @49k'') and where (e.g., ``BMW'')
\end{itemize}

\paragraph{Annotation Span Length \& Methodology:} \begin{enumerate}
\item The annotation span under the Historic can be considered as the most elaborate amongst the other classes. It would refer to events that happened in the past (e.g., ``Replaced clutch'', ``3 owners'', ``Last service was done 3,000 miles ago'', ``primarily done on the motorway'').
\item Should there be a mention of damage and subsequently a relevant mention of component change, these two spans are separate. Treat each of the aforementioned bullet points as a separate span.
\item If a span mentions a component change that still affects the car today, then it's labelled as Historic (e.g., ``2 new front tyres fitted last May'').
\end{enumerate}

\subsection*{Sales Options}
\textit{Sales Options} involve various aspects related to the selling process. 

\begin{tcolorbox}[colback=gray!5!white, colframe=gray!75!black, title=Definition]
The \textit{SALES OPTIONS} label refer to any word or span of words that indicate something tangible or a service offered during or after the sale apart from the car itself, and enhances the sale proposition. 
\end{tcolorbox}

These include:
\begin{itemize}
    \item Part exchange offers
    \item Number of keys
    \item Delivery \& delivery options
    \item Extended Opening times (i.e. unusual times, late night or weekends)
    \item Warranty available
    \item Financing options
    \item Sale of the car's number plate (or not)
    \item Availability of additional photos from past accidents
    \item Offering test drive
    \item Price negotiability
    \item Any documentation, for example:
        \begin{itemize}
            \item Full service history logbook (FSH)
            \item Providing MOT papers
        \end{itemize}
    \item Valeting/Cleaning
    \item And any other kind of service that could be considered an advantage over other sellers
\end{itemize}

\paragraph{Annotation Span Length \& Methodology:} \begin{enumerate}
\item We would label the noun part of the sale, or any adjective specifying it, but disregarding any conditions. For instance, ``2 year warranty \sout{when financed with Volvo Services * Terms and Conditions apply.}''. 
\item Label spans that add value to the ad, according to your judgement. If something seems like a Sales Option but does not add value in comparison with the rest of the ads, then disregard it. Instances of those are contact details and location.

\end{enumerate}

\section{Not Tracked Named Entities}
\label{sec:appendix:not_tracked}

Throughout our iterations, we documented some key themes from the text part of the advertisements that we identified as recurrent features of the texts but not useful for annotation. As such, we explicitly do not annotate for the following reasons:

\begin{itemize}
    \item ``Logistical costs'': This includes the amount of tax, insurance, whether the vehicle complies with the ULEZ\footnote{The acronym refers to the Ultra Low Emission Zone charge in London, where more polluting vehicles are charged accordingly.}, and how long a full tank lasts. However, this information is redundant as it can be calculated more accurately from the car's specifications, given that the time of purchase, location, and type of use will affect these values.

    \item ``Personal Sentiments'': We found statements denoting emotional distress for selling the car. For example, ``it breaks my heart to see this car go''. We omitted these statements, since they do not add to any objective quantifiable metric.
    \item ``Reason for selling'': Moreover, a few advertisements disclosed a reason for selling. These could refer to an individual buying a newer vehicle, or downsizing to a smaller vehicle. Although one could argue these statements may have some value to the buyer, we omit those since again we believe they do not add to an objective quantifiable metric.
    \item ``Car Identifiables'': In multiple advertisements in the text description, sellers tend to reiterate the car model's brand, model, year, colour and engine size. Nonetheless, for any online car advertisement, these statements are included in the car advert header. Hence, we omitted them.    
    \item ``Invitation for communication \& location'': We would avoid labelling references to contact them, or where they are located.
    \item ``Optional \& Standard features'': Detecting car features, both optional and standard, is of great interest, since we have been told by our industry partner that they can change the price evaluation of a vehicle. Nonetheless, throughout our investigations features can differ from vehicle to vehicle. To make the point clearer, what one manufacturer considers as a feature either a standard or an optional one, may not necessarily be one according to another manufacturer's opinion. Hence, we have another piece of work that targets solely the annotation of features based on a dictionary table, casting it as a contextual NER problem. 
    \item ``Sales Pitch'': In certain instances, particularly among traders rather than individual sellers, there are elaborate descriptions that emphasise the vehicle's perceived suitability and appeal to potential buyers, yet without providing substantive factual information.
    
    \end{itemize}

\section{PoS Exploratory Analysis}
\label{sec:appendix:pos}
Please refer to Table~\ref{tab:pos_percentages} for the most frequent part-of-speech metrics per label.

\begin{table}[htbp]
\small
\centering

\resizebox{\columnwidth}{!}{%
\begin{tabular}{|l|c|c|c|}
\hline
\textbf{PoS Tag} & \textbf{HISTORIC} & \textbf{CONDITION} & \textbf{SALES OPTIONS} \\ \hline
NN & 23.9\% & 23.7\% & 37.7\% \\ 
JJ & 14.9\% & 12.5\% & 10.9\% \\ 
NNP & 14.5\% & 13.1\% & 25.7\% \\ 
CD & 13.4\% & 14.1\% & 10.2\% \\ 
NNS & 12.6\% & 11.7\% & 9.9\% \\ 
IN & 7.5\% & 10.2\% & 2.2\% \\ 
VBN & 5.7\% & - & - \\ 
DT & - & 4.6\% & - \\ 
CC & - & - & 1.8\% \\ \hline
\end{tabular}}
\caption{Top PoS Tag Frequencies as Percentages for Each Label. The detailed PoS tags can be found in \url{https://web.archive.org/web/20190206204307/https://www.clips.uantwerpen.be/pages/mbsp-tags}.}
\label{tab:pos_percentages}
\end{table}

\section{IAA Label Analysis}
\label{sec:app:iaa}
Table~\ref{tab:iaa-scores} reports inter-annotator agreement across all annotator pairs, broken down by entity label and measured using both exact and partial match precision, recall, and F\textsubscript{1} as defined in Section~\ref{sec:exp:evaluation}. Scores are computed over all ordered pairs of the three annotators, where each annotator serves as the reference in turn, yielding six pairwise comparisons in total. The raw counts (\textit{COR}, \textit{INC}, \textit{PAR}, \textit{MIS}, \textit{SPU}) are aggregated across all pairs prior to computing the final F\textsubscript{1}. For exact match, the \textit{PAR} category is excluded, and a span must fully agree on both boundaries and label to be counted as correct.

\begin{table}[h]                                                                    
\centering       
\resizebox{\columnwidth}{!}{%
\begin{tabular}{ll ccc}                                                     
\toprule
\textbf{Label} & \textbf{Match} & \textbf{P} & \textbf{R} & \textbf{F\textsubscript{1}} \\                                                                             
\midrule                                                  
\multirow{2}{*}{\textsc{Condition}}
& Exact   & 75.60 & 75.60 & 75.60 \\
& Partial & 81.64 & 81.64 & 81.64 \\
\midrule
\multirow{2}{*}{\textsc{Sales\_Options}}
& Exact   & 64.88 & 64.88 & 64.88 \\
& Partial & 76.83 & 76.83 & 76.83 \\
\midrule
\multirow{2}{*}{\textsc{Historic}}
& Exact   & 67.28 & 67.28 & 67.28 \\
& Partial & 77.83 & 77.83 & 77.83 \\
\bottomrule
\end{tabular}
}
\caption{Inter-annotator agreement (IAA) across all annotator pairs, reported as exact and partial match precision (P), recall (R), and F\textsubscript{1} (\%) per
entity label. Partial match awards half credit to overlapping spans.}
\label{tab:iaa-scores}
\end{table}

\section{Contextual Mention Types}
\label{sec:appendix:families}
Table~\ref{tab:families} gives one gold span per label for four frequent mention types (Section~\ref{sec:ner-schema}). The counts per label over the whole dataset are: MOT 195 / 23 / 115, service 42 / 139 / 388, tyre 34 / 62 / 5 and mileage 103 / 39 / 37 (\textit{Condition} / \textit{Historic} / \textit{Sales Options}).

\begin{table*}[t]
\centering
\small
\begin{tabular}{llll}
\toprule
\textbf{Mention} & \textbf{Condition} & \textbf{Historic} & \textbf{Sales Options} \\
\midrule
MOT & \textit{``MOT Until October''} & \textit{``never failed an MOT''} & \textit{``6 months MOT''} \\
service & \textit{``next service due September 2025''} & \textit{``serviced every 6 months''} & \textit{``full service history''} \\
tyre & \textit{``good tyres''} & \textit{``Tyres recently fitted''} & \textit{``tyre repair kit''} \\
mileage & \textit{``18,500 miles''} & \textit{``new front discs at 64k miles''} & \textit{``mileage verification''} \\
\bottomrule
\end{tabular}
\caption{The same mention type under each of the three labels: gold spans from the dataset. An MOT can be the certificate the car holds now, a test it passed in the past, or a promise made with the sale.}
\label{tab:families}
\end{table*}

\begin{table*}[t]
\centering
\small
\begin{tabular}{llccccc}
\toprule
\textbf{Model} & \textbf{Label} & \textbf{COR} & \textbf{INC} & \textbf{PAR} & \textbf{MIS} & \textbf{SPU} \\
\midrule
\multirow{3}{*}{DeBERTaV3-base} & Historic & 37.8 & 9.5 & 23.5 & 5.7 & 23.5 \\
 & Sales Options & 55.6 & 1.3 & 17.4 & 7.8 & 17.8 \\
 & Condition & 38.3 & 11.7 & 23.8 & 9.7 & 16.4 \\
\midrule
\multirow{3}{*}{GPT-4o} & Historic & 32.3 & 6.0 & 19.3 & 36.7 & 5.7 \\
 & Sales Options & 49.4 & 1.5 & 13.7 & 27.8 & 7.6 \\
 & Condition & 34.7 & 9.8 & 12.5 & 37.0 & 6.0 \\
\midrule
\multirow{3}{*}{Gemini 1.5 Flash} & Historic & 33.5 & 7.7 & 12.9 & 37.1 & 8.7 \\
 & Sales Options & 47.8 & 1.5 & 11.6 & 31.2 & 7.9 \\
 & Condition & 23.6 & 11.8 & 8.0 & 51.3 & 5.3 \\
\midrule
\multirow{3}{*}{CRF} & Historic & 39.6 & 6.6 & 20.8 & 26.4 & 6.6 \\
 & Sales Options & 55.0 & 1.6 & 16.0 & 18.6 & 8.8 \\
 & Condition & 27.7 & 10.2 & 17.2 & 35.8 & 9.1 \\
\bottomrule
\end{tabular}
\caption{Match types per model and label, as percentages of the model's matches for that label pooled over the three test folds (COR correct, INC incorrect label, PAR partial overlap, MIS missing, SPU spurious).}
\label{tab:error_types_per_label}
\end{table*}

\section{Match Types per Label}
\label{sec:appendix:error_types}
Table~\ref{tab:error_types_per_label} breaks Table~\ref{tab:error_types} down by label. As in Table~\ref{tab:res_per_label}, each label's row restricts the gold spans to that label while predicted spans of other labels remain eligible as incorrect-label matches.

\section{Hyper-Parameters \& LM Versions}
\label{sec:appendix:hyper-params}

{\small
\begin{lstlisting}[language=bash, caption={Encoder transformer hyper-parameter settings used in our experiments tailored for our relatively small number of examples.}, label=lst:hyperparameters]
--num_train_epochs 30
--per_device_train_batch_size 16
--per_device_eval_batch_size 16
--load_best_model_at_end True
--metric_for_best_model eval_loss
--greater_is_better False
--warmup_steps 50
--evaluation_strategy epoch
--save_strategy epoch
--learning_rate 1e-5
\end{lstlisting}
}

{\small
\begin{lstlisting}[language=, caption={LM versions used in our experiments.}, label=lst:llm_versions]
- bert-base-cased
- deberta-v3-base
- gpt-4o-2024-11-20
- gemini-1.5-flash-002
- Qwen/Qwen2.5-72B-Instruct
- meta-llama/Meta-Llama-3.1-70B-Instruct
\end{lstlisting}
}

\raggedbottom
\makeatletter
\setlength\@dblfptop{0pt plus 0pt}
\setlength\@dblfpsep{12pt plus 0pt minus 0pt}
\setlength\@dblfpbot{0pt plus 1fil}
\makeatother
\section{CRF Top Features}
\label{sec:appendix:crf_top}
 Tables~\ref{tab:crf_sales}--\ref{tab:crf_historic} list the ten highest-weighted CRF state features per class.                                      
  The feature name follows \texttt{sklearn-crfsuite} conventions:                                                                                     
  \texttt{token.lower()} is the lowercased current token;
  \texttt{token[-3:]} and \texttt{token[-2:]} are the last three and two characters, respectively;                                                    
  \texttt{shape} is spaCy's orthographic shape, where \texttt{x} denotes a lowercase letter, \texttt{X} an uppercase letter, \texttt{d} a digit, and  
  punctuation is preserved verbatim;
  a \texttt{-1:} or \texttt{+1:} prefix indicates the feature is extracted from the immediately preceding or following token.
  Some values reflect noisy, user-generated advertisement text (e.g.\ \emph{longmot+new} is a
  concatenation of ``long MOT'' and ``new'' without spacing; \emph{bar*brand} is a
  stylistic separator common in listings).

  \begin{table*}[t]
  \centering
  \small
  \begin{tabular}{lll}
  \toprule
  \textbf{Feature} & \textbf{Value} & \textbf{Description} \\
  \midrule
  token.lower()     & reconditioning & Current token is ``reconditioning'' \\
  token.lower()     & securing       & Current token is ``securing'' \\
  token.lower()     & fsh            & Current token is ``fsh'' (Full Service History) \\
  token.lower()     & warranty       & Current token is ``warranty'' \\
  token.lower()     & detailing      & Current token is ``detailing'' \\
  -1:token.lower()  & visual         & Previous token is ``visual'' \\
  token.lower()     & delivery       & Current token is ``delivery'' \\
  token.lower()     & finance        & Current token is ``finance'' \\
  token.lower()     & warranties     & Current token is ``warranties'' \\
  +1:token.lower()  & expose         & Next token is ``expose'' \\
  \bottomrule
  \end{tabular}

    \caption{Top ten most significant CRF features for the \textit{Sales Options} class.}
  \label{tab:crf_sales}
  \end{table*}

  \begin{table*}[t]
  \centering
  \small
  \begin{tabular}{lll}
  \toprule
  \textbf{Feature} & \textbf{Value} & \textbf{Description} \\
  \midrule
  token.lower()     & unregistered   & Current token is ``unregistered'' \\
  -1:token.lower()  & category       & Previous token is ``category'' \\
  token.lower()     & category       & Current token is ``category'' \\
  +1:token.lower()  & but            & Next token is ``but'' \\
  -1:token.lower()  & excellent      & Previous token is ``excellent'' \\
  +1:token.lower()  & \faCheck       & Next token is a tick emoji \\
  +1:token.lower()  & 1x             & Next token is ``1x'' \\
  +1:token.lower()  & ready          & Next token is ``ready'' \\
  shape             & dd,ddd         & Token shape matches a mileage pattern, e.g.\ ``81,446'' \\
  token.lower()     & longmot+new    & Noisy concatenation of ``long MOT'' and ``new'' \\
  \bottomrule
  \end{tabular}
    \caption{Top ten most significant CRF features for the \textit{Condition} class.}
  \label{tab:crf_condition}
  \end{table*}

  \begin{table*}[t]
  \centering
  \small
  \begin{tabular}{lll}
  \toprule
  \textbf{Feature} & \textbf{Value} & \textbf{Description} \\
  \midrule
  -1:token.lower()  & fronts         & Previous token is ``fronts'' \\
  -1:token.lower()  & shaft          & Previous token is ``shaft'' \\
  +1:token.lower()  & refurbished    & Next token is ``refurbished'' \\
  token.lower()     & motorway       & Current token is ``motorway'' \\
  +1:token.lower()  & advisories     & Next token is ``advisories'' \\
  token[-3:]        & NER            & Last 3 characters are ``NER'', e.g.\ from ``OWNER'' \\
  token.lower()     & advisories     & Current token is ``advisories'' \\
  token.lower()     & refurbished    & Current token is ``refurbished'' \\
  token.lower()     & bar*brand      & Stylistic separator in ad text \\
  shape             & xxx*xxxx       & Token shape of the above (lowercase*lowercase) \\
  \bottomrule
  \end{tabular}
    \caption{Top ten most significant CRF features for the \textit{Historic} class.}
  \label{tab:crf_historic}
  \end{table*}

\end{document}